\documentclass[conference]{IEEEtran}
\IEEEoverridecommandlockouts
\usepackage{cite}
\usepackage{amsmath,amssymb,amsfonts}
\usepackage{algorithmic}
\usepackage{graphicx}
\usepackage{textcomp}
\usepackage{xcolor}
\usepackage{flushend,cuted}
\usepackage{geometry}
\usepackage{caption}
\usepackage{amsmath, amssymb}
\usepackage{makecell}
\usepackage{multirow}
\usepackage{array}
\usepackage{longtable}
\usepackage{bbding}
\usepackage{amssymb}
\usepackage{diagbox}
\usepackage{booktabs}
\usepackage{graphicx} 
\usepackage{float} 
\usepackage{subfigure} 
\usepackage{CJK}
\def\BibTeX{{\rm B\kern-.05em{\sc i\kern-.025em b}\kern-.08em
    T\kern-.1667em\lower.7ex\hbox{E}\kern-.125emX}}
\begin{document}
\hyphenpenalty=5000
\tolerance=1000
\title{Parallel Multi-Graph Convolution Network For Metro Passenger Volume Prediction\\
}

\author{\IEEEauthorblockN{Fuchen Gao \textsuperscript{1} }
\IEEEauthorblockA{College of Information Science and \\
Engineering \\
East China University of Science and \\
Technology\\
Shanghai, China\\
y30190782@outlook.com}
\and
\IEEEauthorblockN{Zhanquan Wang \textsuperscript{2,*} }
\IEEEauthorblockA{College of Information Science and\\
 Engineering \\
East China University of Science and\\
 Technology\\
Shanghai, China\\
zhqwang@ecust.edu.cn}
\and
\IEEEauthorblockN{Zhenguang Liu \textsuperscript{3} }
\IEEEauthorblockA{College of Computer Science and\\
 Technology\\
Zhejiang University\\
Zhejiang, China \\
liuzhenguang2008@gmail.com}}

\maketitle

\begin{abstract}
Accurate prediction of metro passenger volume (number of passengers) is valuable to realize real-time metro system management, which is a pivotal yet challenging task in intelligent transportation. Due to the complex spatial correlation and temporal variation of urban subway ridership behavior, deep learning has been widely used to capture non-linear spatial-temporal dependencies. Unfortunately, the current deep learning methods only adopt graph convolutional network as a component to model spatial relationship, without making full use of the different spatial correlation patterns between stations. In order to further improve the accuracy of metro passenger volume prediction, a deep learning model composed of Parallel multi-graph convolution and stacked Bidirectional unidirectional Gated Recurrent Unit (PB-GRU) was proposed in this paper. The parallel multi-graph convolution captures the origin-destination (OD) distribution and similar flow pattern between the metro stations, while bidirectional gated recurrent unit considers the passenger volume sequence in forward and backward directions and learns complex temporal features. Extensive experiments on two real-world datasets of subway passenger flow show the efficacy of the model. Surprisingly, compared with the existing methods, PB-GRU achieves much lower prediction error. \\
\end{abstract}


\section{Introduction}
Traffic prediction is one of the fundamental tasks of urban traffic management, which provides necessary information for intelligent transportation applications such as travel route planning and travel demand assessment \cite{b1}. With the rapid expansion of cities, metro plays an improving role in the urban public transportation system, and the prediction of metro ridership has gradually become a hot topic.

In recent years, the development of deep learning provides new paradigms for traffic forecasting, which promotes the vigorous progress of this area \cite{b2}. In particular, Graph Convolutional Network (GCN) provides a more feasible way for modeling spatial dependencies in traffic networks \cite{b10}. Given the inherent graph structure of the transportation network, GCN is able to preserve the real topology and capture the dependencies between metro stations. However, the effective construction of the graph and the structure of the GCN network are still two important problems remain to be solved. For the first issue, previous work directly uses the physical topology to build the graph \cite{b3}. However, besides the physical adjacency relationship, two stations may also have other semantic correlations. For example, two stations may share similar passenger volume patterns or have stable passenger flows between them. Considering these facts, in this paper additionally utilizes two useful station connections which are shown in the Fig. 1. More specifically, they are: 

\textbf{Flow Pattern Similarity:} Intuitively, two different stations in the city that belong to the same functional area (e.g.  residential area) may have the same passenger flow pattern.

\textbf{Origin-Destination Flow Direction:} The Origin-Destination (OD) distribution of ridership represents the correlation between two stations. For example, if most of the passenger flow at a residential area station flows into a commercial area station, the two stations are related. The connection between them is undoubtedly valuable for future passenger volume prediction. However, using only the physical topology of the metro system would unfortunately miss this useful information.
\begin{figure}[H] 
\centering 
\vspace{-0.4cm} 
 \captionsetup{labelformat=default,labelsep=period,font={small}}
\setlength{\abovecaptionskip}{0.2cm}   
 \setlength{\belowcaptionskip}{0cm}  
 \captionsetup{labelformat=default,labelsep=period,font={small}}
\includegraphics[width=8cm]{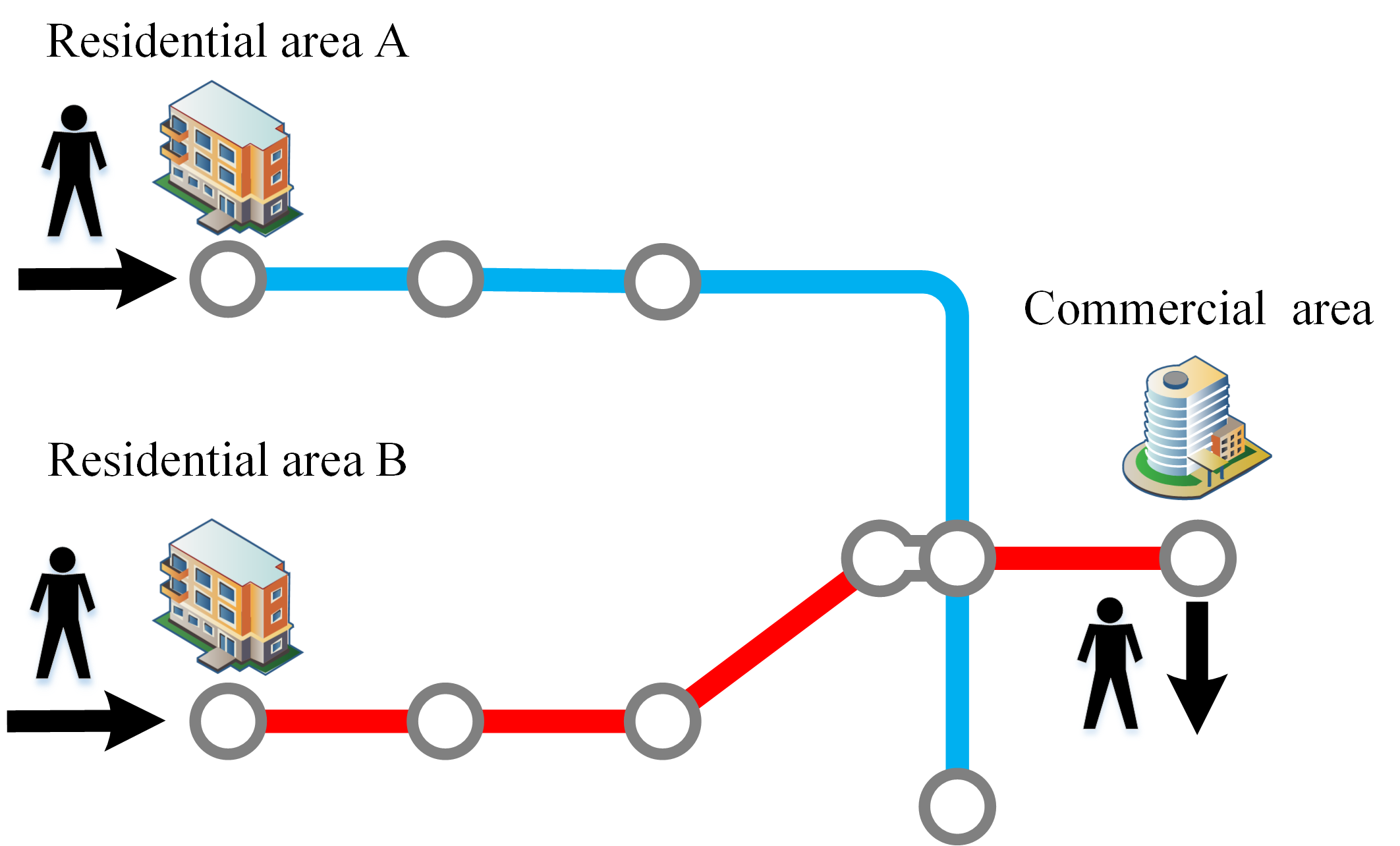} 
\caption{The spatial correlation } 
\label{Fig.1} 
\end{figure}
\vspace{-0.3cm} 
On the other hand, there is still no appropriate graph convolutional network that could elegantly incorporate such three kinds of semantic correlations, namely physical adjacency, flow pattern similarity, and origin-destination flow direction. This motivates us to proposes Parallel multi-graph convolution and stacked Bidirectional unidirectional gated recurrent unit model (PB-GRU) for metro passenger volume prediction. In PB-GRU, two graph convolution modules with different structures, FSGCN and FDGCN, are constructed to model the Flow Pattern Similarity and OD Flow Direction respectively. The Stacked bidirectional unidirectional temporal Attention Gated Recurrent Unit (SAGRU) is used to deal with time series and is capable of capturing long-term temporal dependencies. This network structure focuses on modeling different spatial correlations in the metro system simultaneously, and decouples the prediction of inflow and outflow series to reduce network complexity. In summary, the key contributions can be summarized as follows:

\begin{itemize}
\item We propose a new deep learning model termed PB-GRU for spatial-temporal representation learning. It incorporates three predefined graph and historical passenger flow series for passenger volume prediction.\\
\item We design different graph convolution networks to capture Flow pattern similarity and OD flow direction between metro stations. We introduce  Simple Graph Convolution (SGC) to decouple the message passing mechanism from GCN. This allows the graph convolution modules to incorporate custom transform learning network for better performance.\\
\item Extensive experiments on two real-world metro ridership benchmark datasets show that our method has much improved performance compared to the existing methods on station-level passenger volume prediction. The ablation studies further verify the effectiveness of the two proposed graph convolution modules.\\
\end{itemize}
\section{REALATED WORK}
Deep learning's achievements in computer vision \cite{b29} and natural language processing have made it widely used in traffic prediction\cite{b14,b15}. In early works, traffic data was directly used to train a Recurrent Neural Network (RNN) for learning long or short-term dependencies \cite{b19}. Cui et al. studied the performance of stacked LSTM in traffic prediction and obtained promising results by using pure RNN structure \cite{b20}. Nonetheless, RNN still has inherent shortcomings in capturing spatial relationships, which led the researchers to introduce CNN as a spatial module. Shi et al. embedded convolution into the gate mechanism and built Conv-LSTM to solve precipitation nowcasting \cite{b22}. Zhang et al. designed an end-to-end residual convolution network to predict the crowds in every city region \cite{b21}.

In fact, traditional convolutional neural network can only be applied for Euclidean data. \cite{b17,b18} However , the graph convolution generalizes the traditional convolution to non-Euclidean structure data. In recent years, the development of graph-neural networks with different structures has promoted the research of spatial correlations in traffic prediction \cite{b23,b24,b25}. Graph convolution method consists of spectral-based and spatial-based methods. The spectral-based methods process the graph signal by introducing filters, while the spatial methods directly aggregating feature information from node’s neighbors. Li et al. \cite{b4} proposed to model the spatial dependence of traffic as a diffusion process, and designed the DCRNN model with Encode-Decode structure by integrating GRU and spectrum-based GCN. Yu et al. \cite{b5} used Gated Linear Unit Convolution to replace RNN and selected Chebyshev convolution operator to construct a pure convolutional structure to extract spatial-temporal features. Wu et al. \cite{b6}proposed a method to generate the adaptive adjacent matrix. This spatial-based approach can learn spatial associations without predefined graphs. However, the adjacent matrix obtained by node embedding only learned the overall similarity of nodes in the training set, resulting in the risk of over-fitting. Liu et al. \cite{b7} incorporated the physical topology, ridership similarity, and inter-station ridership into a Graph Convolution Gated Recurrent Unit. However, due to the large number of parameters in PVGCN, the model is hard to be trained. Thanks to the research on parallel deep learning structure\cite{b8}, Chen et al. \cite{b9} proposed a parallel GCN-SUBLSTM framework for subway passenger volume prediction. The results show that this parallel structure can significantly improve the accuracy of prediction and has high training efficiency. Nevertheless, GCN-SUBLSTM lacks the design of graph convolutional network structure. Therefore , it is unable to completely surpass PVGCN in accuracy.

To address these limitations, this paper proposes a parallel multi-graph convolution and stacked Bidirectional Unidirectional Gated Recurrent Unit improving the GCN-SUBLSTM. After carefully investigating the recent research about graph convolution structure\cite{b26,b27,b28}, we use SGC as graph convolution operation, \cite{b11} and design different network structures for transform learning. This enables our model to obtain lower prediction error.

\section{Preliminaries}
\subsection{Problem Definition}\label{AA}
The inflow and outflow volumes of station i at time interval t are denoted as a two-dimensional vector $x_i^t$ $\in$ $\mathbb{R}^2$, where the first element represents inflow volume while the second element captures outflow volume. The whole metro system at time interval t is represented as ($x_1^t$$,$$x_2^t$$,$$\cdots$$,$$x_n^t$) $\in$ $\mathbb{R}$$^{2\times n}$, where n is the number of stations. Assume future volume series is (\textit{X$^{t+1}$$,$X$^{t+2}$$,$$\cdots$$,$X$^{t+T}$}) $\in$ $\mathbb{R}$$^{2\times n\times T}$, our model can be viewed as learning a mapping function \textit{f}($\cdot$) that:
\begin{equation}
\left(X^{1}, X^{2}, \cdots, X^{\mathrm{t}}\right) \stackrel{\mathrm{f}(\cdot)}{\Rightarrow}\left(X^{\mathrm{t}+1}, X^{\mathrm{t}+2}, \cdots, X^{\mathrm{t}+\mathrm{T}}\right)
\end{equation}
\subsection{Graph Definition Units}\label{BB}
\noindent \textbf{Multi-hop Physical Graph.} It is widely recognized that there is a close ridership relationship between two stations that are physically adjacent to each other. Therefore, this paper designs a multi-hop physical graph to describe the spatial structure of the subway network. Given a positive integer K, if station i only can reach station j by a minimum of K edges, we set $A^{(k)}_{ij}$ to 1.
\begin{equation}
A_{i j}^{(K)}=\left\{\begin{array}{l}
1, \partial=K \\
0, \text { otherwise }
\end{array}\right.
\end{equation}
where $\partial$ represents the minimum number of edges required to connect from station i and station j. When K=1, the first-order physical graph can be regarded as a special case of multi-hop graph, which does not affect the realization of traditional graph convolution. The advantage of multi-hop physical graph is that it extends the representation ability of traditional graph convolution.\\
\textbf{Passenger Volume Pattern Similarity Graph.} If the passenger volume curves of two metro stations are similar, the two stations may have similar functions in the physical world and similar passenger volume patterns. Considering the order of magnitude difference of passenger volume in different metro stations, Dynamic Time Warping (DTW) is engaged for constructing thus share similar graphs.\cite{b16} According to the definition of the DTW algorithm, the passenger volume pattern similarity matrix is given by:
\begin{equation}
S_{i j}=\exp \left(D T W\left(x_{i} x_{j}\right)\right)
\end{equation}
where  $x_i$ and $x_j$  represent the historical sequence of the node i and j. \textit{S} is the passenger volume pattern similarity matrix obtained through DTW algorithm. Finally, Top-k and threshold method are used to further filter out the edges with smaller values to avoid the matrix being too dense.\\
\textbf{OD Flow Direction Graph.} Ridership direction is also an important of correlation between metro stations, reflecting the regular daily migration of urban population. The model uses the origin and destination distribution of ridership to construct the OD flow direction graph. Let F($\textit{i, j}$) be the total number of passengers from station j to station i in the whole training set. The weight $C_{ij}$  is calculated as follow:
\begin{equation}
C_{i j}=\frac{F(i, j)}{\sum_{n=1}^{N} F(i, n)}
\end{equation}
where \textit{N} represents the number of nodes with passenger flow to station i, and \textit{C} represents the overall OD flow direction matrix. Since the sum of each row in the matrix is always less than or equal to 1, only some edges with very small weight will be removed to eliminate the noise.\\
\textbf{Multi-hop Degree Matrix.}  Graph regularization can effectively improve the performance of graph neural network in feature extraction. In this section, a new multi-hop degree matrix is proposed to achieve more flexible and reasonable graph regularization for the multi-hop random diffusion process of passenger flow in the metro system. The multi-hop matrix can be expressed as:
\begin{equation}
D^{(k)}=\left\{\begin{array}{l}
d\left(A_{i j}^{(k)}\right), k=1 \\
D^{(k-1)}+d\left(A_{i j}^{(k)}\right), \text { otherwise }
\end{array}\right.
\end{equation}
where $d(\cdot)$  represents the operation of the degree matrix, k stands for the hops of the physical graph, and $D^k$ is a unique diagonal matrix. Note that the adjacency matrix $A^{(k)}_{ij}$ of the physical graph is defined as a symmetric matrix.
\section{The PB-GRU Model}
Our model mainly consists of three modules: Stacked bidirectional unidirectional temporal Attention Gated Recurrent Unit (SAGRU), Flow Similarity Graph Convolution Network (FSGCN), and Flow Direction Graph Convolution Network (FDGCN). SAGRU is used to learn the complex time dependency from the historical passenger flow series. Considering the differences in graph structure and semantics, two completely different structures are designed for the graph convolution modules to better extract the spatial and temporal correlations between the station nodes. Finally, the output embeddings of all modules are passed through the dropout layer, and two fully connected layers to predict the inflow and outflow volumes respectively.
\subsection{Stacked Bidirectional Unidirectional Temporal Attention \indent Gated Recurrent Unit}
As a variant of LSTM, GRU has been widely used for time series modeling. We would like to point out that the single-layer GRU can only capture the positive dependencies within the time series due to the presence of reset gates, which inevitably loses useful information in the implicit states when new inputs are absorbed.

Bidirectional GRU (Bi-GRU) can help address this issue by building a two-directional GRU layer. It uses hidden states from both directions to compensate for the information loss in forward propagation. Therefore, for time series prediction, the Bi-GRU has better ability to capture long-term dependencies and make more accurate predictions. As shown in Fig. 2, the bidirectional GRU contains two parallel GRU layers in forward and backward directions respectively. In SAGRU, the output of bidirectional GRU can be formulated as:
\begin{equation}
\vec h_t = GRU_{forward}(X^t,\vec h_{t-1})
\end{equation}
\begin{equation}
\vec h_t = GRU_{backward}(X^t,\overleftarrow h_{t+1})
\end{equation}
\begin{equation}
b_t = \vec h_t + \overleftarrow h_t
\end{equation}
where $GRU_{forward}$ and $GRU_{backward}$ represent the forward and backward GRU respectively, and $\vec{h_t}$ and  $\overleftarrow{h_t}$ are two hidden states learned from the Bi-GRU. $b_t$  represents the output of each time step. Bidirectional hidden states can be fused in the form of concatenation or addition. The addition function is chosen in the model, which can effectively reduce the number of parameters in the next layer.

The predictive ability of neural networks can be improved by deepening the model structure. Stacked Unidirectional Bidirectional Recurrent Neural Network (SUBRNN) has been shown to be able to generate higher level feature representations from time series \cite{b13}. Therefore, this study constructed SAGRU to learn the temporal dependence of passenger volume sequences. As Fig. 2, the output $H_t$ of the bidirectional GRU is fed into the stacked GRU layer for higher-level representation learning. The output of the stacked forward GRU is given as:
\begin{equation}
u_t = GRU_{stack}(b_t,u_{t-1})
\end{equation}
where $GRU_{stack}$ represents the stacking GRU, and $u_{t-1}$ is the output of time step \textit{t}. In existing methods, vector  $u_{t}$ can be directly fed into a decoding network for prediction, but SAGRU additionally adopts a lightweight temporal attention to assign more weights to important features in the fusion tensor. Formally,
\begin{equation}
\begin{array}{c}
e_{t}=\tanh \left(W_{u} u_{t}+b_{u}\right) 
\end{array}
\end{equation}
\begin{equation}
\begin{array}{c}
a_{t}=\frac{\exp \left(e_{t}\right)}{\sum_{t=1}^{H} \exp \left(e_{t}\right)} 
\end{array}
\end{equation}
\begin{equation}
\begin{array}{c}
O_{GRU}=\sum\limits_{t=1}^{H} a_{t} u_{t}
\end{array}
\end{equation}
\begin{figure*}[htbp] 
\centering 
 \captionsetup{labelformat=default,labelsep=period,font={small}}
\setlength{\abovecaptionskip}{0.2cm}   
 \setlength{\belowcaptionskip}{0cm}  
\includegraphics[width=15cm]{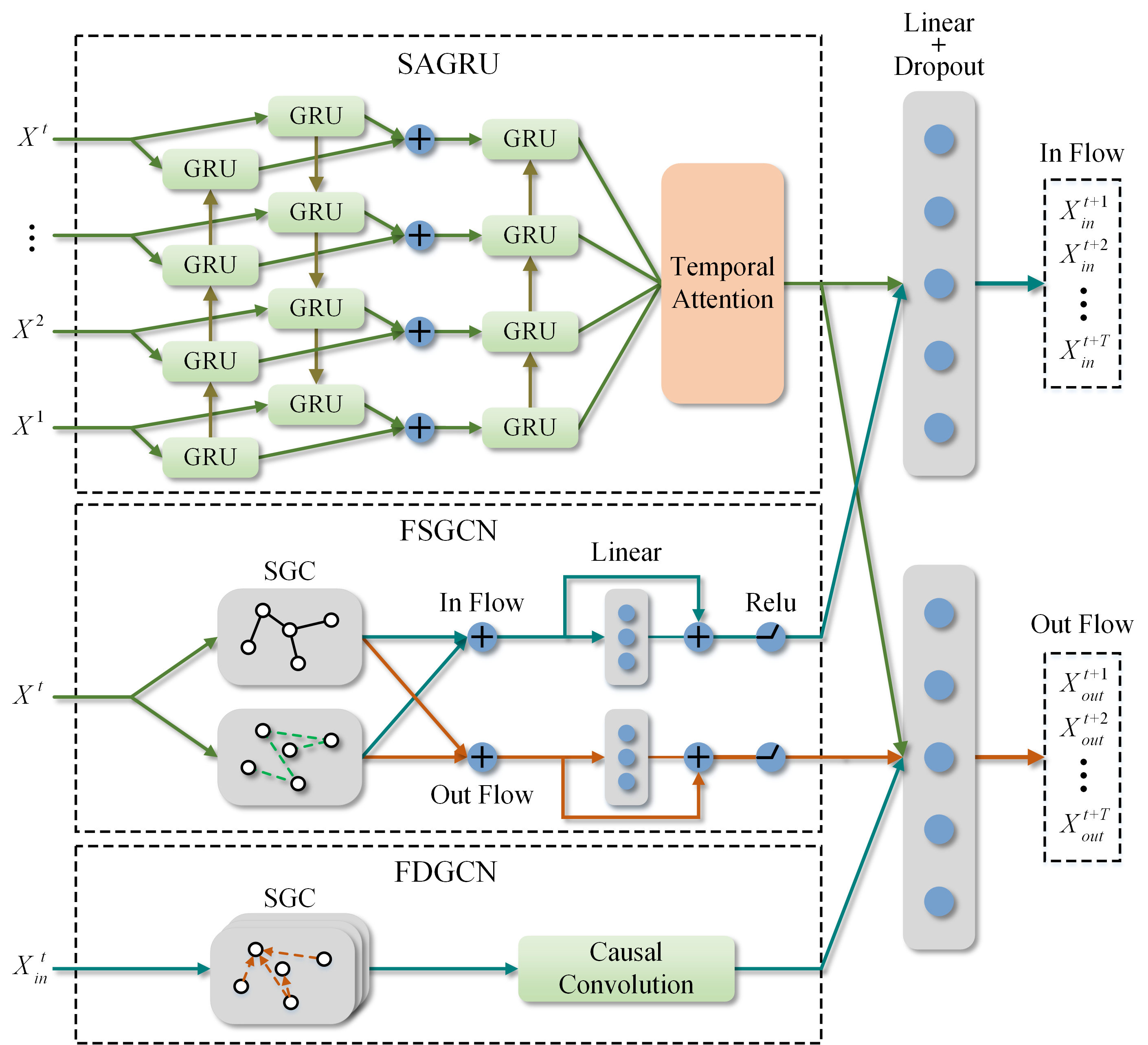} 
\caption{The overall architecture  of PB-GRU } 
\label{Fig.2} 
\end{figure*}
\noindent where $W_u$, $b_u$ represent weight and bias parameters respectively, and $a_t$ stands for the weight of attention after normalization. Finally, the hidden states $u_t$ are weighted and added to obtain the output representation $O_{RGU}$ of SAGRU.
\subsection{Flow similarity Graph Convolution Network}
GCN has become a new and effective mean in spatial modeling. However, it was originally designed for graph classification tasks and is equipped with a number of neural network operations. Existing studies have pointed out that the most common feature transformation and nonlinear activation operations in GCN contribute little to some tasks and make the model difficult to train [12]. 

In this paper, SGC is introduced as a graph convolution operation. Its advantage is that SGC only uses the message passing mechanism to aggregate the original graph signals and avoids the multi-channel linear transformation. The FSGCN module uses first-order physical graph and ridership pattern similarity graph to capture the similarity of ridership patterns.
\begin{equation}
\begin{array}{c}
H_{P}=\left(D^{-1 / 2} A_{i j} D^{-1 / 2}\right) X^{t} 
\end{array}
\end{equation}
\begin{equation}
\begin{array}{c}
H_{S}=\left(D^{-1} S\right) X^{t}
\end{array}
\end{equation}
where \textit{D} is the first-order degree matrix, $A_{ij}$ is the first-order physical adjacency matrix, and \textit{S} is the ridership pattern similarity graph. According to the difference of prior knowledge, Laplace regularization is used for physical graphs and random walk regularization is used for similar graphs. It has been proven that message passing and feature transformation of graph neural network can be completely decoupled [11],[26]. Our model also adopts a residual linear layer at the end. In order to avoid overfitting,  $H_{p}$ and $H_{s}$ are separated after SGC and concatenate as inflow embedding $H_{if}$  and outflow embedding $H_{of}$.
\begin{equation}
\begin{aligned}
O_{FSif} &=\operatorname{Relu}\left(H_{if}+\left(W_{if} H_{if}+b_{if}\right)\right)
\end{aligned}
\end{equation}
\begin{equation}
\begin{aligned}
O_{FSof} &=\operatorname{Relu}\left(H_{of}+\left(W_{of} H_{if}+b_{of}\right)\right)
\end{aligned}
\end{equation}
where $W_{if}$, $b_{if}$, $W_{if}$ , $b_{if}$ are all trainable parameters. The nonlinear residual layer realizes the decoupling of feature transformation and message passing. It is designed to capture the potential node relationship better.
\subsection{Flow Direction Graph Convolution Network}
Due to the lack of prior knowledge of passenger flow, FDGCN models the passenger flow as a multi-hop random diffusion process. Given a constant K without considering few passengers coming in and out of the same station, the model assumes that the passengers are equally likely to arrive at all stations within K hop, and builds a group of K-hop flow direction graphs to aggregate fine-grained hierarchical flow information.
\begin{equation}
\begin{aligned}
h^{(K)}=\left(\left(A^{(K)}\odot C\right) D^{(K)^{-1}}\right) X_{if}
\end{aligned}
\end{equation}
where $A^{(k)}$ represents the k-hop physical graph, C represents the OD Flow Direction graph, $D^{(k)}$ represents the K-hop degree matrix, $\odot$  represents the Hadamard product, and \textit{K} is a hyper parameter. We denoted the this set of graph convolution results as $h_{hid}$ , which represents the potential passenger flow between subway stations in \textit{k} hop.

For a station in the metro system, the farther the station, the later the arrival time. The K-hop potential passenger flow can be regarded as time series, and due to the predicted time interval, the beneficial flow for prediction should be a subset of $h_{hid}$. Next, a causal convolution is applied to further extract the potential passenger flow, as shown in Fig. 3 Before each convolution layer, zero padding was applied. Thus, the output has the same length as the input, and the network can only use the information of past time steps. In addition, gate mechanism is also used in causal convolution:
\begin{equation}
\begin{aligned}
P=w_{1} * h_{h i d}+b_{1}
\end{aligned}
\end{equation}
\begin{equation}
\begin{aligned}
Q=w_{2} * h_{h i d}+b_{2}
\end{aligned}
\end{equation}
\begin{equation}
\begin{aligned}
h_{o u t}=\tanh \left(P+h_{h i d}\right) \odot \operatorname{sigmoid}(Q)
\end{aligned}
\end{equation}
P and Q are the results of input $h_{hid}$ through different convolution kernels $w_{1}$ and $w_{2}$, parameters  $b_{1}$ and $b_{2}$ are bias, and $\odot$ represents Hadamard product. By stacking a small number of causal convolution layers, the model can obtain the appropriate size of the receptive field to capture the local continuous passenger flow. The final output is represented as $O_{FDof}$ after being cut.
\begin{figure} [H]
\centering 
 \captionsetup{labelformat=default,labelsep=period,font={small}}
\setlength{\abovecaptionskip}{0.2cm}   
 \setlength{\belowcaptionskip}{0cm}  
\includegraphics[width=8cm]{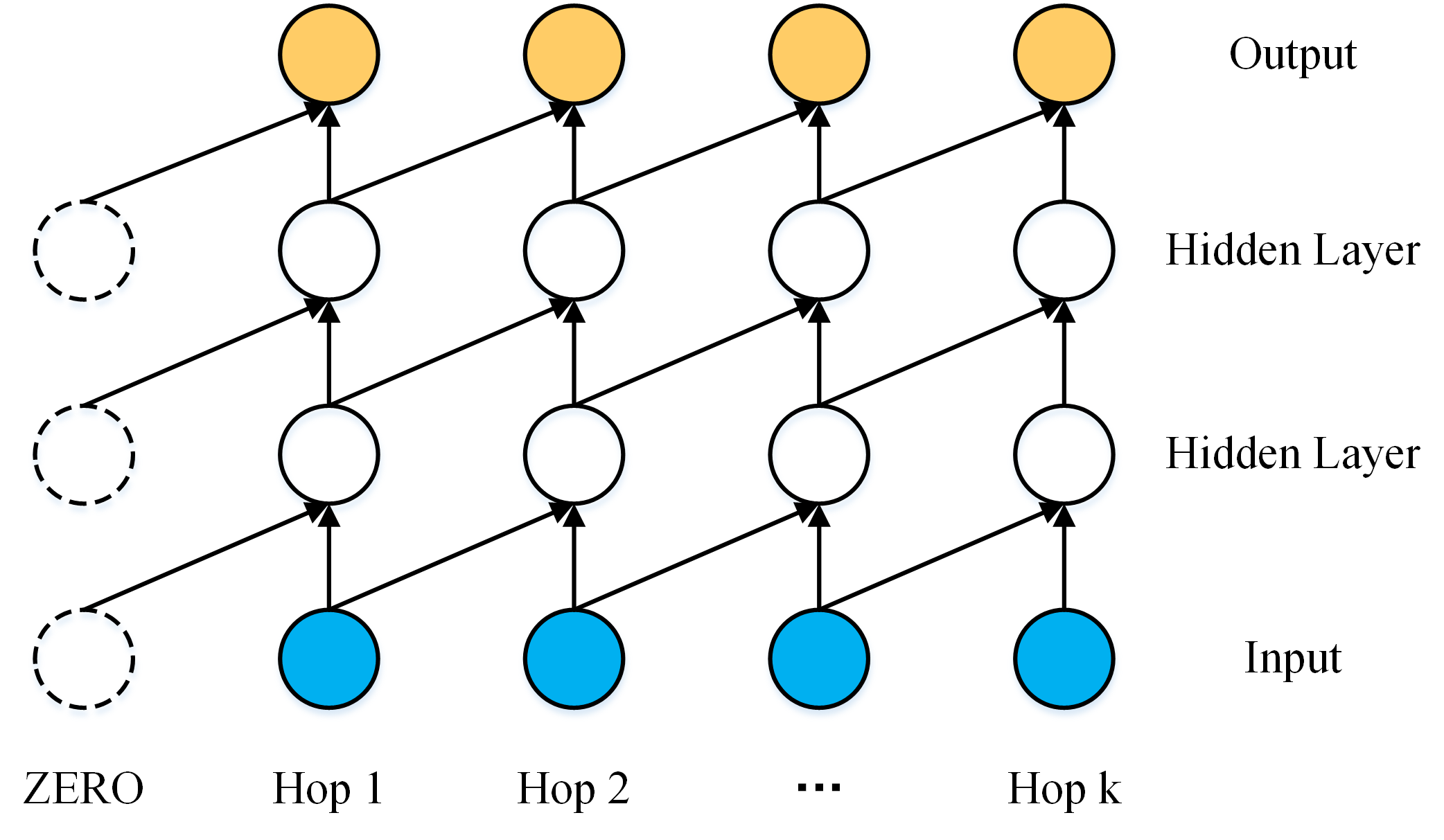} 
\caption{Causal Convolution} 
\label{Fig.3} 
\end{figure}
\subsection{Output layer}
Finally, the model uses two independent fully connected layers to fuse the embeddings in order to predict the future inflow and outflow volumes respectively. It can obtain the predicted value of T time steps at one time.
\begin{equation}
\begin{aligned}
\hat{Y}^{\text {in }}=W_{\text {in }}\left(O_{G R U} \| O_{F S i f}\right)+b_{i n}
\end{aligned}
\end{equation}
\begin{equation}
\begin{aligned}
\hat{Y}^{\text {out }}=W_{\text {out }}\left(O_{G R U}\left\|O_{F S o f}\right\| O_{F D o f}\right)+b_{\text {out }}
\end{aligned}
\end{equation}
where $W_{in}$, $W_{out}$, $b_{in}$,  $b_{out}$ represent the trainable weight and bias parameters, $\lVert$ represents the concatenate operation, $\hat{Y}^{\text {in }}$ and $\hat{Y}^{\text {out}}$ are the inflow and outflow of all subway stations respectively. Finally, $\hat{Y}^{\text {in }}$ and $\hat{Y}^{\text {out}}$ are stacked to $\hat{Y}$ as the predicted output of our model.
\section{Experiments}
\subsection{Data sets}
Two metro ridership benchmark datasets, HZMetro and SHMetro, are used to evaluate the performance of our method. The details of these two datasets are summarized in Table I. The training set, verification set, and test set are divided according to the proportion of the original source \cite{b7}, and only Z-score normalization is used for preprocessing.
\subsection{Parameter Settings and Evaluation Metrics}
\textbf{Hyper-parameters:} When training on HZMetro, the batchsize is set to 48, the hidden state size of the GRU is 650. As for SHMetro, batchsize and hidden state size are 96 and 1,200 respectively. The ADAM optimizer is used to minimize the L1 loss to train 350 epochs. The initial learning rate is set at $10^{-3}$ and decays to 0.5 per 40 epochs. The regularization method is Dropout $\textit{p}$ = 0.4,  for SAGRU and FDGCN, and $\textit{p}$ = 0.1 for others.

\textbf{Evaluation Metrics:} Average absolute error (MAE), mean absolute percentage error (MAPE) and root mean square error (RMSE) are used to evaluate the performance of different methods. The three indexes are defined as follows:
\begin{equation}
\begin{array}{l}
M A E=\frac{1}{N} \sum\limits_{i=1}^{N}\left|\hat{y}^{i}-y^{i}\right|
\end{array}
\end{equation}
\begin{equation}
\begin{array}{l}
M A P E=\frac{1}{N} \sum\limits_{i=1}^{N} \frac{\left|\hat{y}^{i}-y^{i}\right|}{y^{i}}
\end{array}
\end{equation}
\begin{equation}
\begin{array}{l}
R M S E=\frac{1}{N} \sum\limits_{i=1}^{N} \frac{\left|\hat{y}^{i}-y^{i}\right|}{y^{i}}
\end{array}
\end{equation}
where $y^i$  represents the ground-truth , $\hat y^i$ represents the predicted value, and \textit{N} represents the number of metro station.\\
\begin{table}[H]
\setlength{\tabcolsep}{8mm}{
\renewcommand{\arraystretch}{1.5}

\caption{DATA SETS}
\label{tab1}
\centering
\begin{tabular}{ccc}
\hline
Dataset& HZMetro& SHMetro\\
\hline
City&\makecell[c]{HangZhou,\\China}& \makecell[c]{ShangHai, \\China}\\
\hline
Station&80&288\\
\hline
Physical Edge&248&958\\
\hline
Ridership/Day&2.35 million&8.82 million\\
\hline
Time Interval&15min&15min\\
\hline
Training Set&\makecell[l]{1/01/2019 – \\ 1/18/2019}&\makecell[l]{7/01/2016 –\\ 8/31/2016}\\
\hline
Validation Set&\makecell[l]{1/19/2019 –\\ 1/20/2019}&\makecell[l]{9/01/2016 – \\ 9/09/2016}\\
\hline
Testing Set&\makecell[l]{1/21/2019 – \\ 1/25/2019}&\makecell[l]{9/10/2016 – \\ 9/30/2016}\\
\hline
\end{tabular}}
\end{table}
\begin{table*}
 \captionsetup{labelformat=default,labelsep=period}
	\caption{EXPERIMENTAL RESULTS ON HZMetro}
	\centering
	\renewcommand{\arraystretch}{1.2}
	\resizebox{0.83\textwidth}{!}{
		\begin{tabular}[htbp]{cccccccccc} 
		\toprule
		\multirow{2}*{Model}&\multicolumn{3}{c}{15min}&\multicolumn{3}{c}{30min}&\multicolumn{3}{c}{60min}\\
		\cline{2-10}
		&MAE&MAPE&RMSE&MAE&MAPE&RMSE&MAE&MAPE&RMSE\\
		\hline
		LSTM&23.43&14.41&40.13&24.38&15.54&42.33&26.74&19.88&47.90\\
		GCN&24.21&21.05&49.20&25.75&24.26&52.34&33.44&27.62&62.39\\
		DCRNN&23.24&13.65&41.43&25.78&15.32&43.23&27.15&18.64&48.56\\
		STGCN&23.86&12.48&45.03&26.07&13.72&49.16&31.47&16.95&59.74\\
		Graph-Wavenet&23.50&13.77&41.88&24.75&15.68&43.70&27.85&20.45&48.69\\
		PVGCN&22.20&\textbf{13.15}&38.12&23.13&13.87&40.00&24.55&16.35&42.26\\
		GCN-SUBLSTM&22.22&13.16&39.83&\textbf{22.84}&13.76&41.08&24.58&15.73&44.48\\
		PB-GRU&\textbf{22.13}&13.30&\textbf{36.55}&22.90&\textbf{13.75}&\textbf{38.33}&\textbf{23.91}&\textbf{14.87}&\textbf{40.02}\\
		\bottomrule
	\end{tabular}}
\label{table_detection results}
\end{table*}
\begin{table*}
 \captionsetup{labelformat=default,labelsep=period}
	\caption{EXPERIMENTAL RESULTS ON SHMETRO}
	\centering
	\renewcommand{\arraystretch}{1.25}
	\resizebox{0.83\textwidth}{!}{
		\begin{tabular}[htbp]{cccccccccc} 
		\toprule
		\multirow{2}*{Model}&\multicolumn{3}{c}{15min}&\multicolumn{3}{c}{30min}&\multicolumn{3}{c}{60min}\\
		\cline{2-10}
		&MAE&MAPE&RMSE&MAE&MAPE&RMSE&MAE&MAPE&RMSE\\
		\hline
		LSTM&23.50&20.23&47.08&24.50&22.64&49.63&26.87&26.16&56.53\\
		GCN&24.21&21.05&49.20&25.75&24.26&52.34&31.60&34.25&63.24\\
		DCRNN&23.34&18.02&47.24&25.33&19.12&51.31&29.01&21.52&63.32\\
		STGCN&23.84&18.71&47.18&26.99&19.41&57.40&33.82&23.69&77.00\\
		Graph-Wavenet&23.75&20.23&45.73&27.12&21.42&54.15&31.56&24.92&68.10\\
		PVGCN&22.85&16.95&45.47&24.16&18.83&50.18&26.37&19.67&58.49\\
		GCN-SUBLSTM&22.75&16.50&46.09&23.77&17.62&49.04&25.87&20.21&55.41\\
		PB-GRU&\textbf{22.70}&\textbf{15.98}&\textbf{43.35}&\textbf{23.74}&\textbf{16.52}&\textbf{46.46}&\textbf{25.72}&\textbf{17.90}&\textbf{51.80}\\
		\bottomrule
	\end{tabular}}
\label{table_detection results}
\end{table*}
\begin{table*}
 \captionsetup{labelformat=default,labelsep=period}
	\caption{EXPERIMENTAL RESULTS OF FSGCN VARIANTS ON HZMETRO}
	\centering
	\renewcommand{\arraystretch}{1.2}
	\resizebox{0.81\textwidth}{!}{
		\begin{tabular}[htbp]{cccccccccc} 
		\toprule
		\multirow{2}*{Model}&\multicolumn{3}{c}{15min}&\multicolumn{3}{c}{30min}&\multicolumn{3}{c}{60min}\\
		\cline{2-10}
		&MAE&MAPE&RMSE&MAE&MAPE&RMSE&MAE&MAPE&RMSE\\
		\hline
		BASE&23.19&13.86&38.43&23	.75&13.94&42.01&25.23&15.86&43.12\\
		P-BASE&22.31&13.49&37.12&23.20&13.79&39.42&24.01&14.82&40.66\\
		D-BASE&22.45&13.58&37.41&23.23&13.82&39.05&24.17&14.83&40.96\\
		PD-BASE&22.21&13.42&37.01&22.96&13.76&38.51&23.94&14.71&40.47\\
		\bottomrule
	\end{tabular}}
\label{table_detection results}
\end{table*}
\begin{table*}
 \captionsetup{labelformat=default,labelsep=period}
	\caption{ EXPERIMENTAL RESULTS OF FSGCN VARIANTS ON SHMETRO}
	\centering
	\renewcommand{\arraystretch}{1.2}
	\resizebox{0.81\textwidth}{!}{
		\begin{tabular}[htbp]{cccccccccc} 
		\toprule
		\multirow{2}*{Model}&\multicolumn{3}{c}{15min}&\multicolumn{3}{c}{30min}&\multicolumn{3}{c}{60min}\\
		\cline{2-10}
		&MAE&MAPE&RMSE&MAE&MAPE&RMSE&MAE&MAPE&RMSE\\
		\hline
		BASE&23.46&16.25&45.78&24	.30&17.72&48.46&26.18&19.61&55.14\\
		P-BASE&23.00&16.03&44.04&23.97&16.50&46.98&26.06&17.77&53.31\\
		D-BASE&23.02&16.03&44.21&24.07&16.51&47.36&26.25&17.70&54.17\\
		PD-BASE&22.80&15.96&43.76&23.79&16.42&46.86&25.79&17.50&52.05\\
		\bottomrule
	\end{tabular}}
\label{table_detection results}
\end{table*}
\subsection{Compared Models}
We compared our method with a variety of existing models including:

\textbf{LSTM:} Long-short term memory neural network, this paper uses encoder - decoder structure to achieve prediction.

\textbf{GCN:} Graph convolution neural network, implemented according to \cite{b10}.

\textbf{DCRNN \cite{b4}:} The diffusion convolutional recurrent neural network uses the bidirectional random walk on the graph combined with the encoder-decoder structure to learn the spatial-temporal dependence.

\textbf{STGCN \cite{b5}:} Spatial and Temporal Graph Convolutional Network, using Chebyshev Graph Convolution and GLU to capture spatial-temporal dependence respectively.

\textbf{Graph-Wavenet \cite{b6}:} The method proposes an adaptive dependency matrix to capture hidden spatial correlations and uses stacked causal convolution to process time series. 

\textbf{PVGCN \cite{b7}:} The physical virtual graph convolutional network combines multi-graph convolution with GRU's gating mechanism, and uses encode-decoder structure to achieve multi-step prediction.

\textbf{GCN-SUBLSTM\cite{b9}:} This model constructs a composite graph, which uses double-layer multi-channel general GCN to capture the spatial correlation of metro stations.

\newpage
\subsection{Performance Comparison}
Performance of different models on HZMetro and SHMetro datasets are shown in Table II and Table III. For all the models, only LSTM and GCN capture a single temporal or spatial correlation. The GCN model performs the worst since it completely ignores the modeling of the time dimension. The LSTM achieves competitive results, but its MAPE is higher than the other models. This empirical evidence suggests that LSTM has poor fitting ability for medium and low passenger flow stations. STGCN and Graph-Wavenet are both time convolution model with Graph convolution, but MAE and RMSE on the two data sets are inferior to LSTM. This indicates that the temporal convolution model is not suitable for short sequences with long time intervals. Both PVGCN and GCN-SUBLSTM have relatively accuracy. PVGCN has a better prediction result on HZMetro, while GCN-SUBLSTM has a greater advantage in the long-term prediction of RMSE index in SHMetro. However, these two models did not use independent graph convolution method to model different spatial-temporal correlation patterns of subway passenger flow. Our PB-GRU achieves the best accuracy in almost all indicators, especially in SHMetro with complex subway network and large passenger flow, the prediction performance is better than all other models. Experimental results show that RMSE of the model is significantly improved compared with the suboptimal model. This proves that the model is more sensitive to the sudden changes of passenger volume and can better predict the marginal values.

\subsection{Ablation Study}
In order to study the contribution of each component in our method, we further conduct extensive experiments by removing each individual component respectively. In particular, we tried the following structures and conducted performance comparison. 

\textbf{Base:} Only SAGRU was used for the prediction.

\textbf{P-Base:} FSGCN with first-order physical graph and SAGRU were used for prediction.

\textbf{D-Base:} FSGCN with ridership pattern similarity graph and SAGRU were used for prediction.

\textbf{PD-Base:} Complete FSGCN and SAGRU were used for co-prediction.
\begin{figure}[H] 
\centering 
 \captionsetup{labelformat=default,labelsep=period,font={small}}
\setlength{\abovecaptionskip}{0.2cm}   
 \setlength{\belowcaptionskip}{0cm}  
\includegraphics[width=8cm]{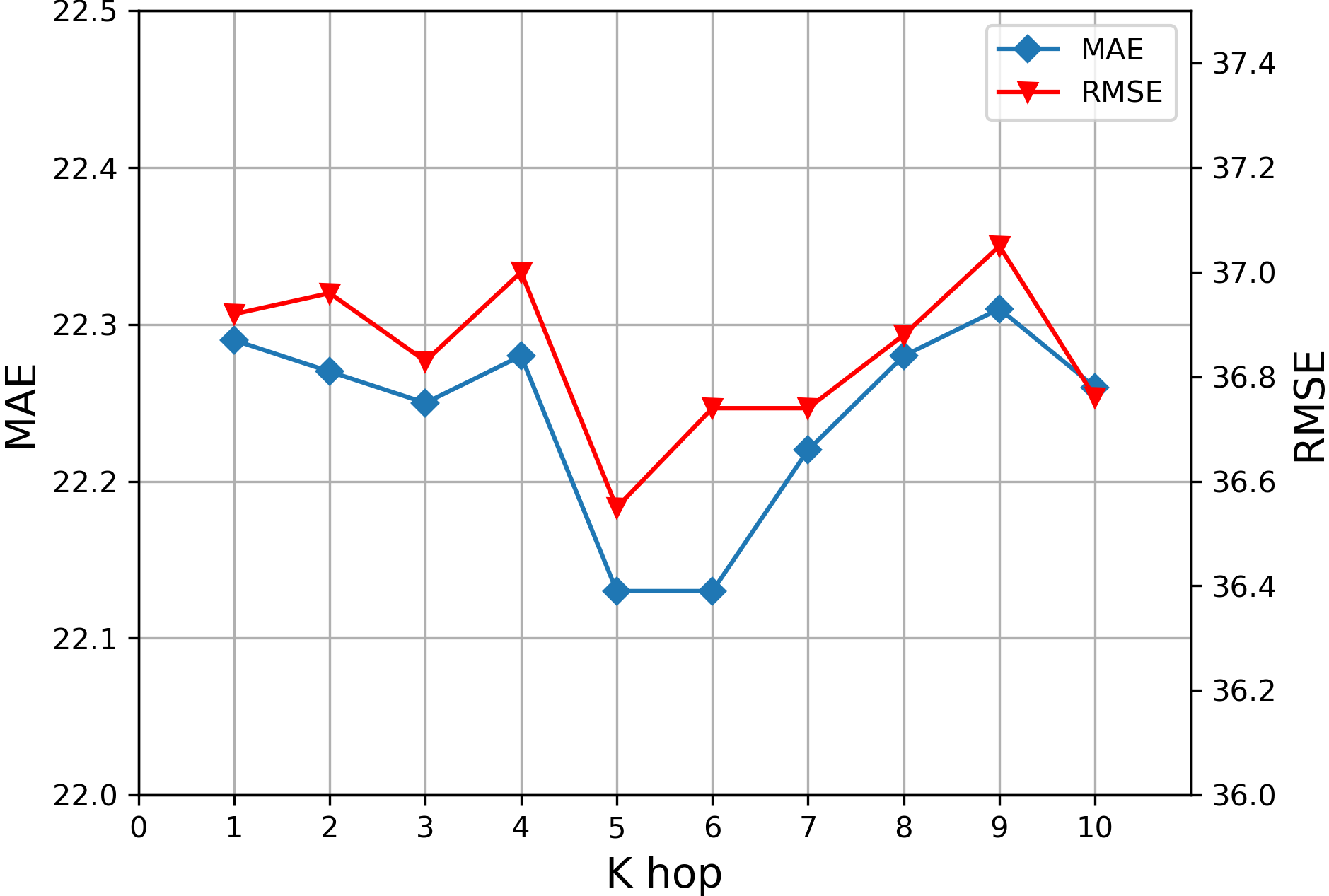} 
\caption{Influence of hyper-parameter K on HZMetro } 
\label{Fig.4} 
\end{figure} 
\begin{figure}[H] 
\centering 
 \captionsetup{labelformat=default,labelsep=period,font={small}}
\setlength{\abovecaptionskip}{0.2cm}   
 \setlength{\belowcaptionskip}{0cm}  
\includegraphics[width=8cm]{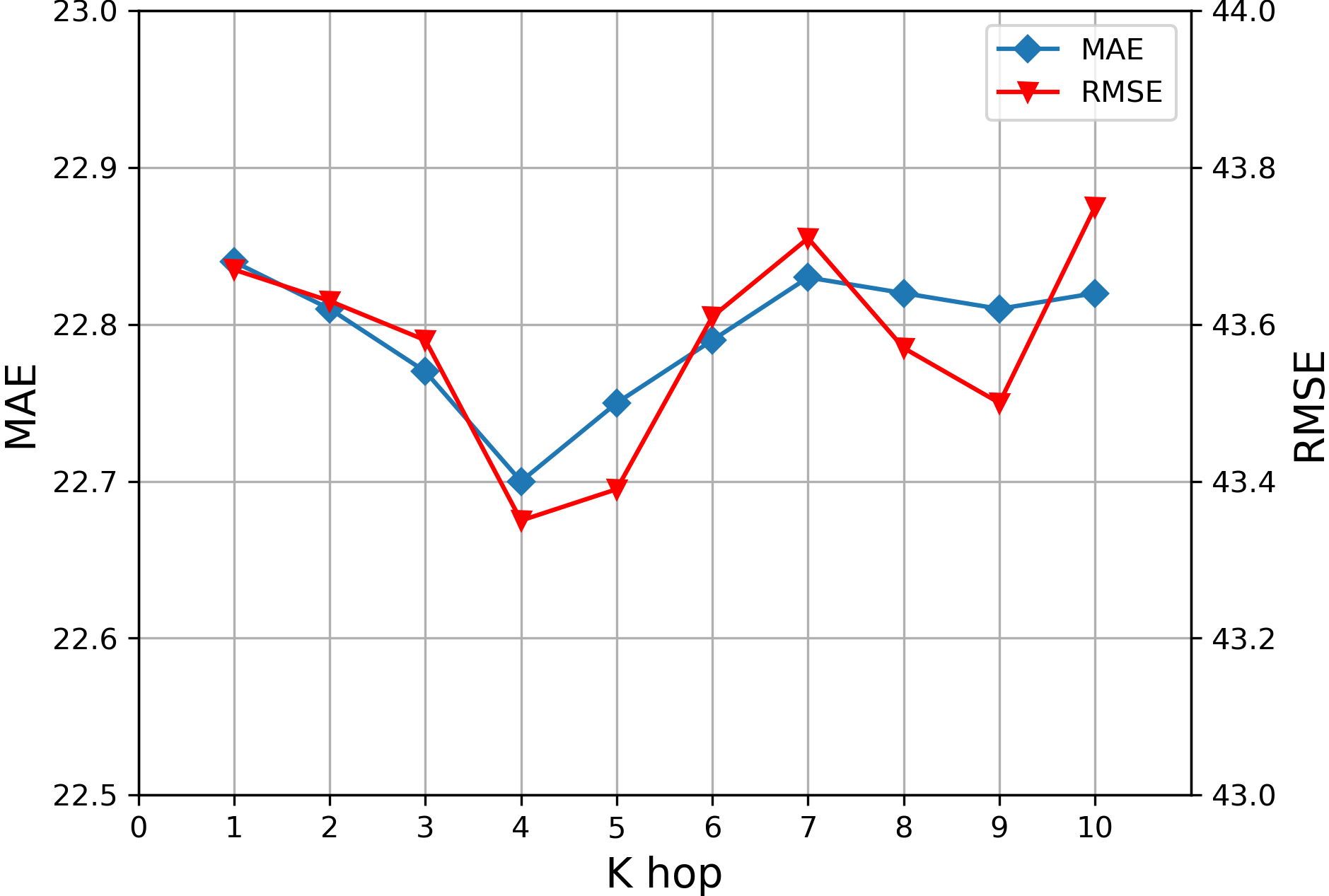} 
\caption{Influence of hyper-parameter K on SHMetro} 
\label{Fig.5} 
\end{figure}
The results are shown in Table IV and Table V. It can be seen from the tables that FSGCN , which is constructed based on the similarity of traffic patterns between stations, significantly reduces the prediction error. As the predicted time interval becomes longer, the RMSE of PD-Base grows more gently than that of Base, which proves that FSGCN captures the similar patterns between metro stations. When only a single similar graph is used for convolution, the accuracy of P-base is higher than D-base, especially on SHMetro. In capturing similar traffic patterns, the first-order physical graph is stronger than the similar graph constructed by DTW, which means that the traffic patterns of adjacent subway stations in big cities are closely correlated. Finally, PD-Base obtained the result with lower error by integrating the embeddings from two kinds of graph convolution, which is competitive with PB-GRU to a certain extent.
\begin{figure}[H] 
\centering 
 \captionsetup{labelformat=default,labelsep=period,font={small}}
\setlength{\abovecaptionskip}{0.2cm}   
 \setlength{\belowcaptionskip}{0cm}  
\includegraphics[width=8cm]{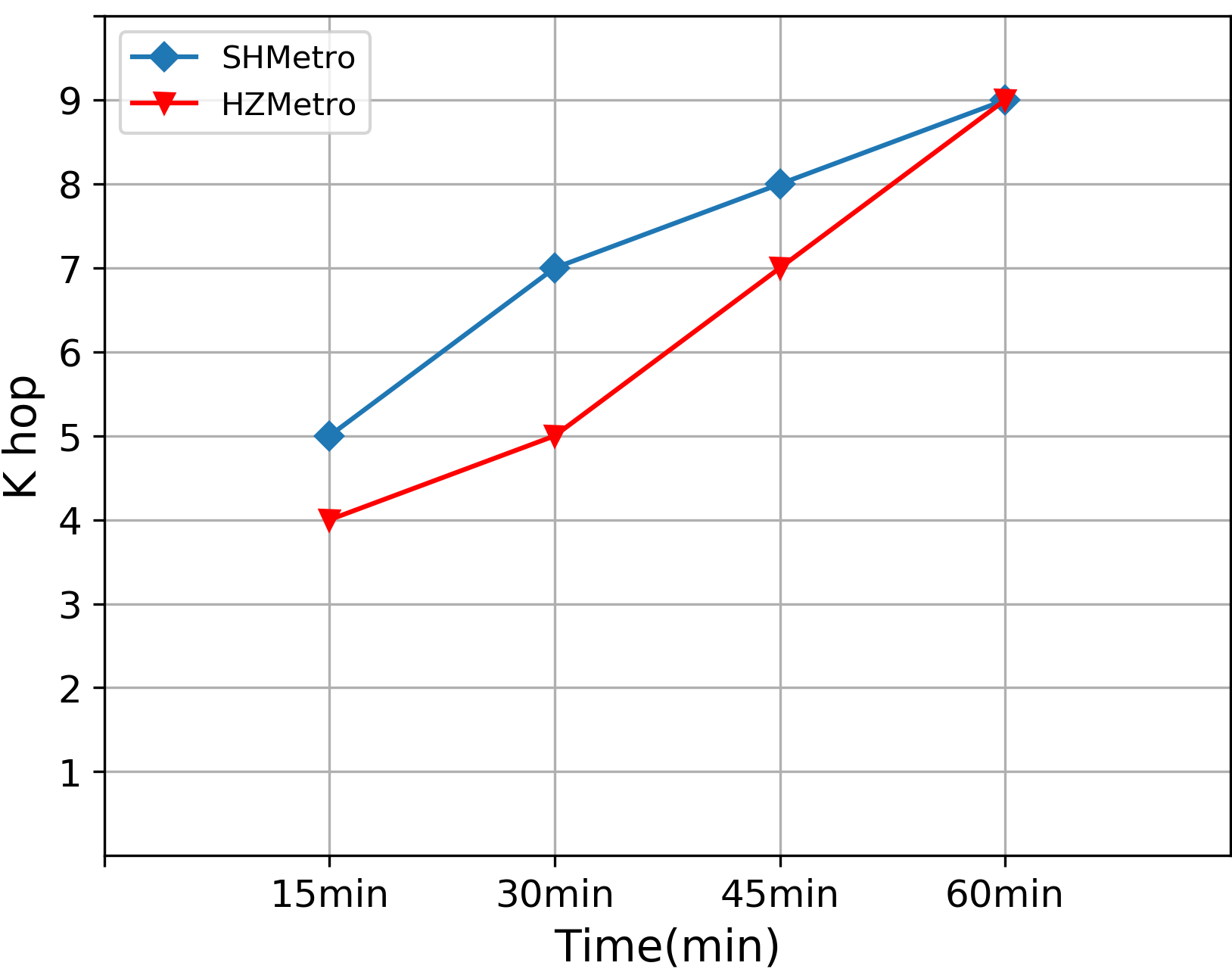} 
\caption{The optimal value of K for different prediction intervals} 
\label{Fig.6} 
\end{figure}
In order to explore the influence of different K-hop hyper-parameter in FDGCN, Fig. 4 and Fig. 5 show MAE and RMSE values of different K values ranging from 1 to 10 in 15min prediction task . In the table, RMSE and MAE start at a high value, then gradually decrease to a minimum value, and finally increase as K become larger. In the predicted time interval of 15min, the optimal K value of Shanghai Metro and Hangzhou Metro was set to 4 and 5, respectively. Fig. 6 shows the best hyper-parameter setting when the minimum
\begin{figure*}[h]
 \captionsetup{labelformat=default,labelsep=period,font={small}}
\centering
\subfigure{
\includegraphics[width=8cm]{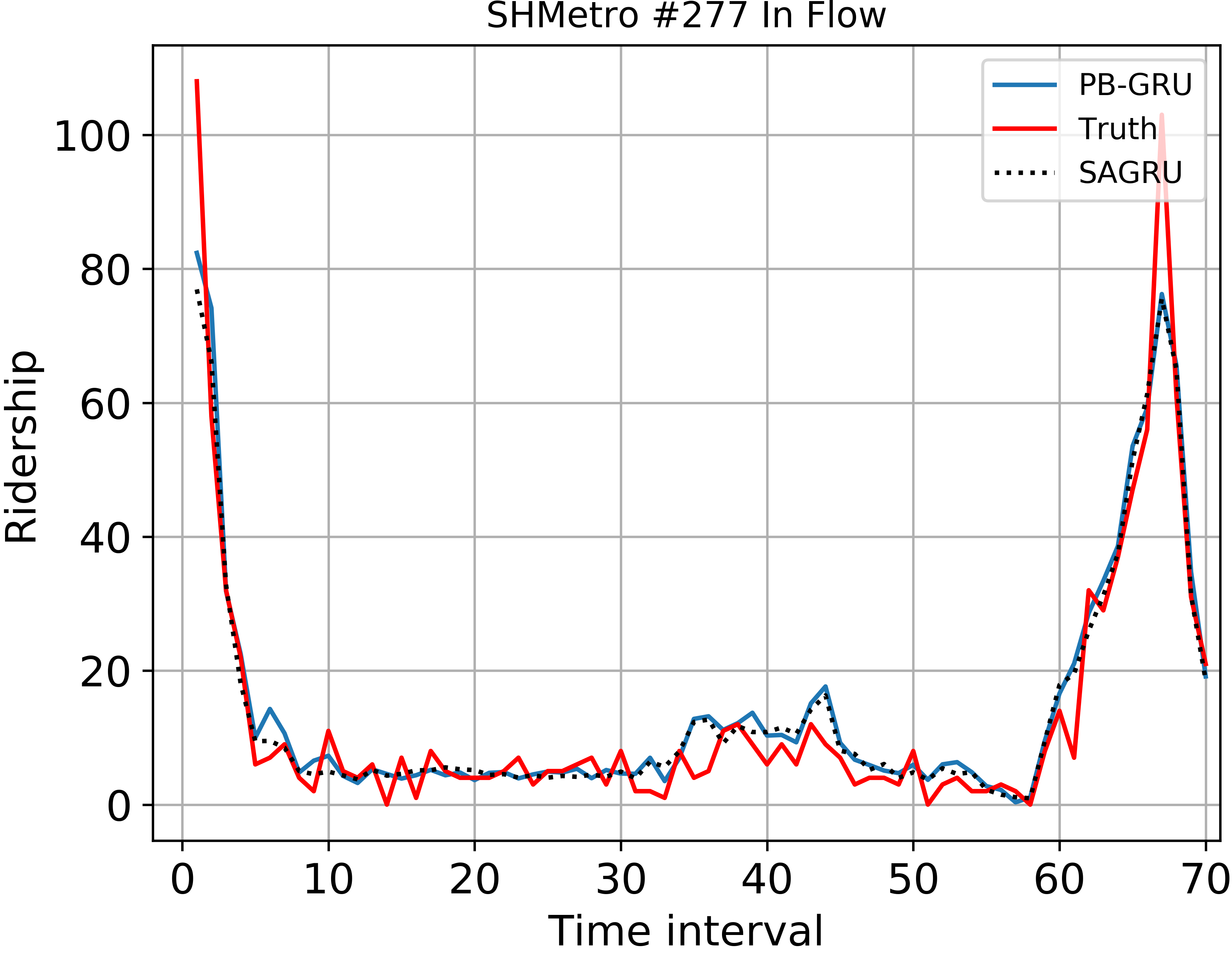}
}
\quad
\subfigure{
\includegraphics[width=8cm]{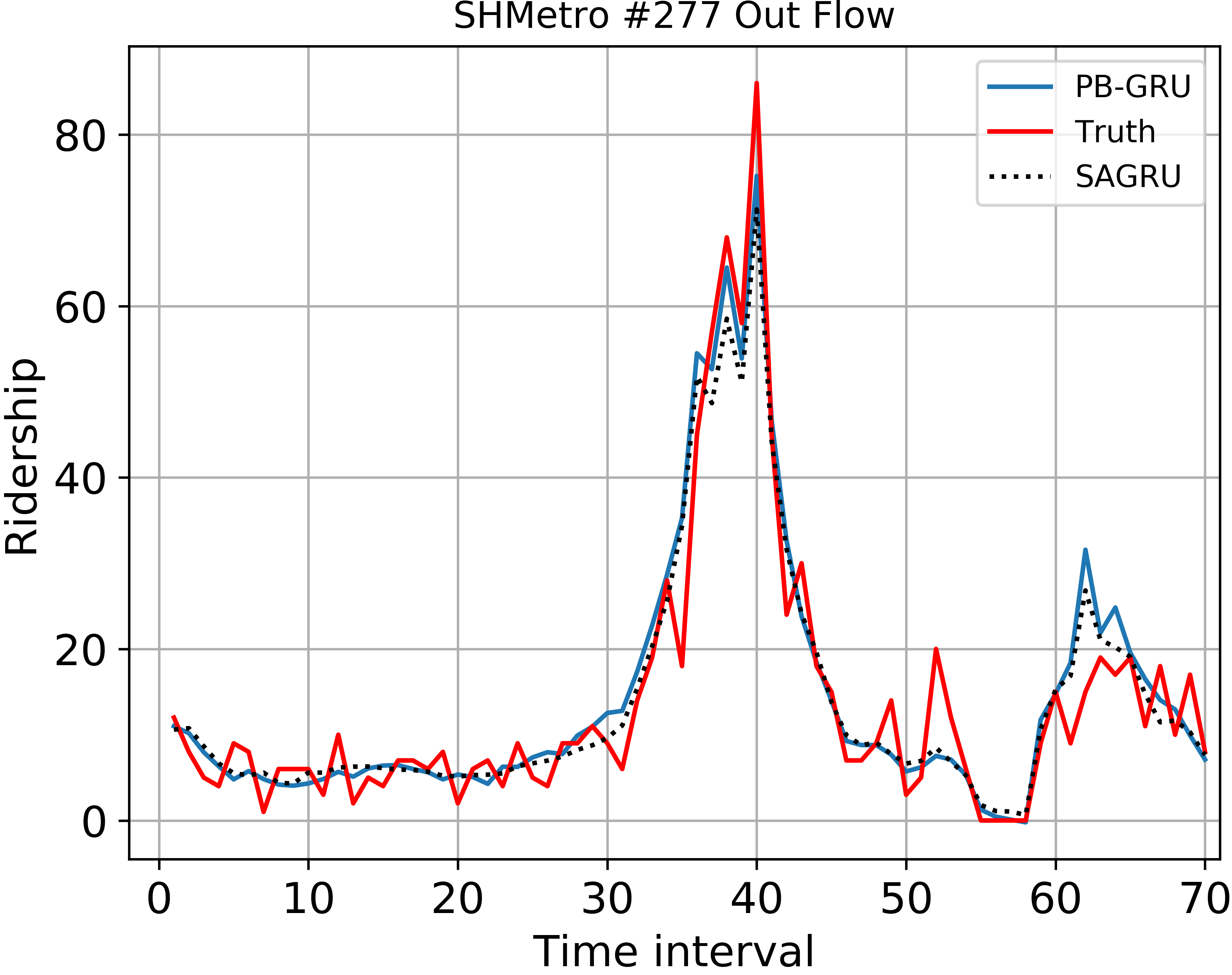}
}
\quad
\subfigure{
\includegraphics[width=8cm]{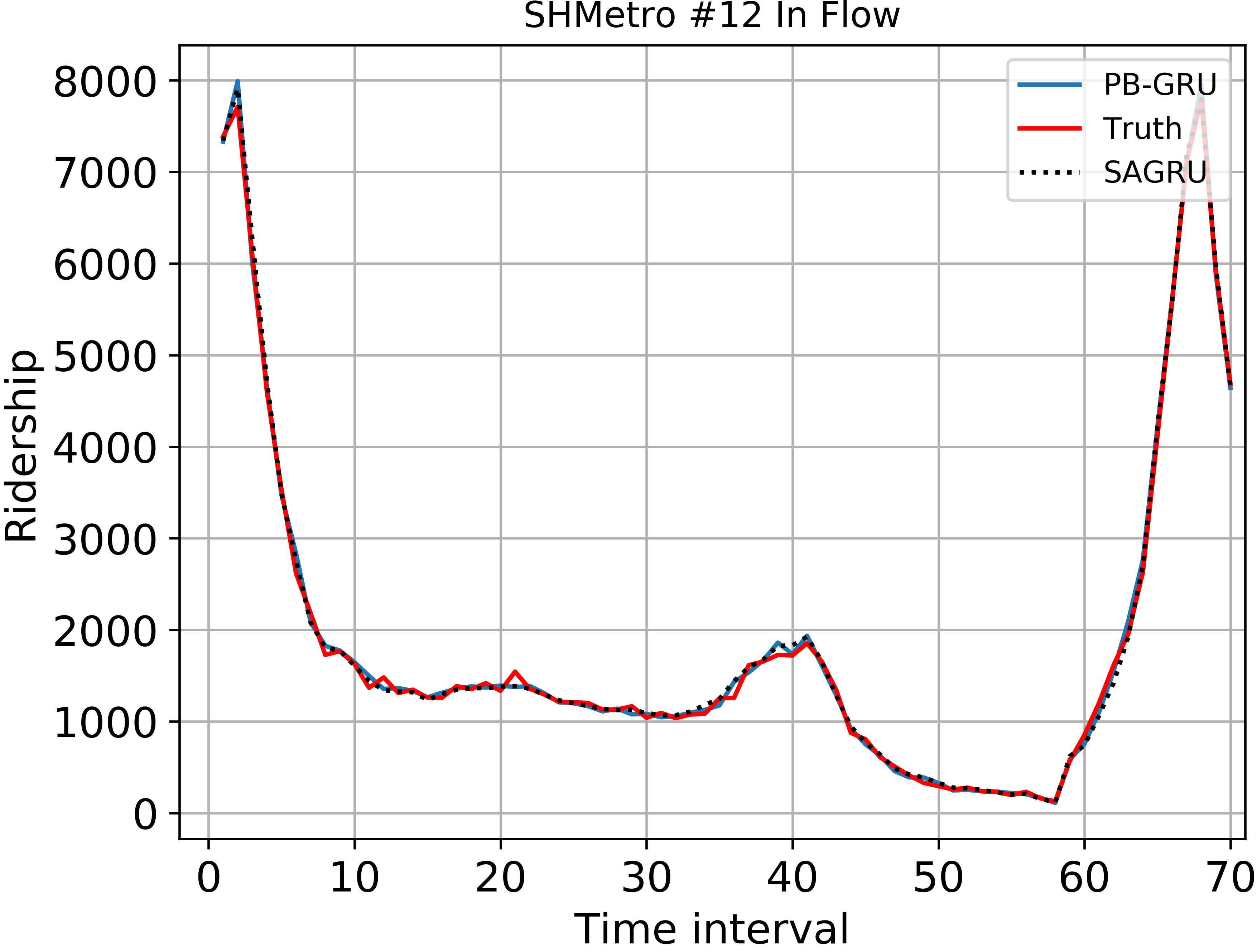}
}
\quad
\subfigure{
\includegraphics[width=8cm]{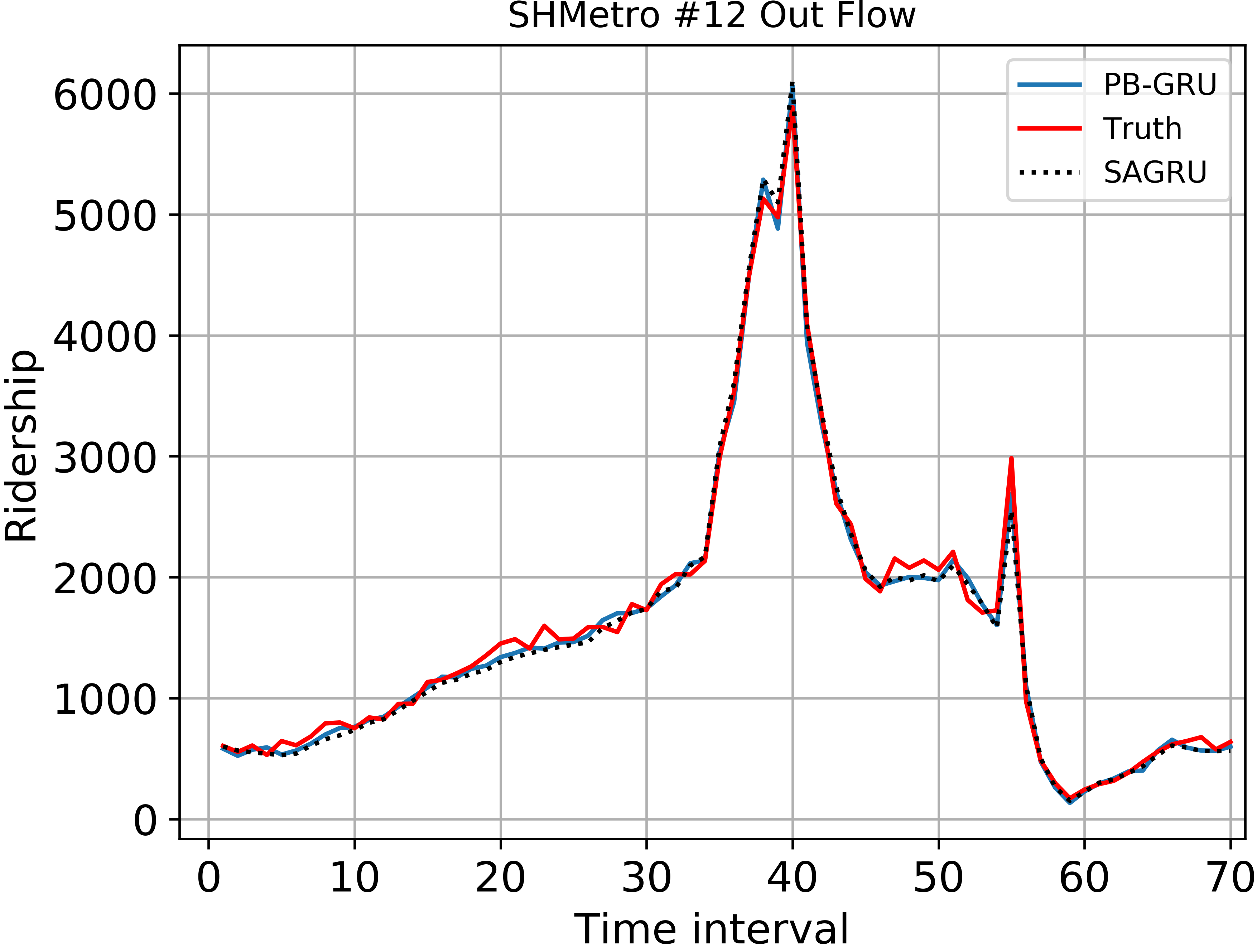}
}
\subfigure{
\includegraphics[width=8cm]{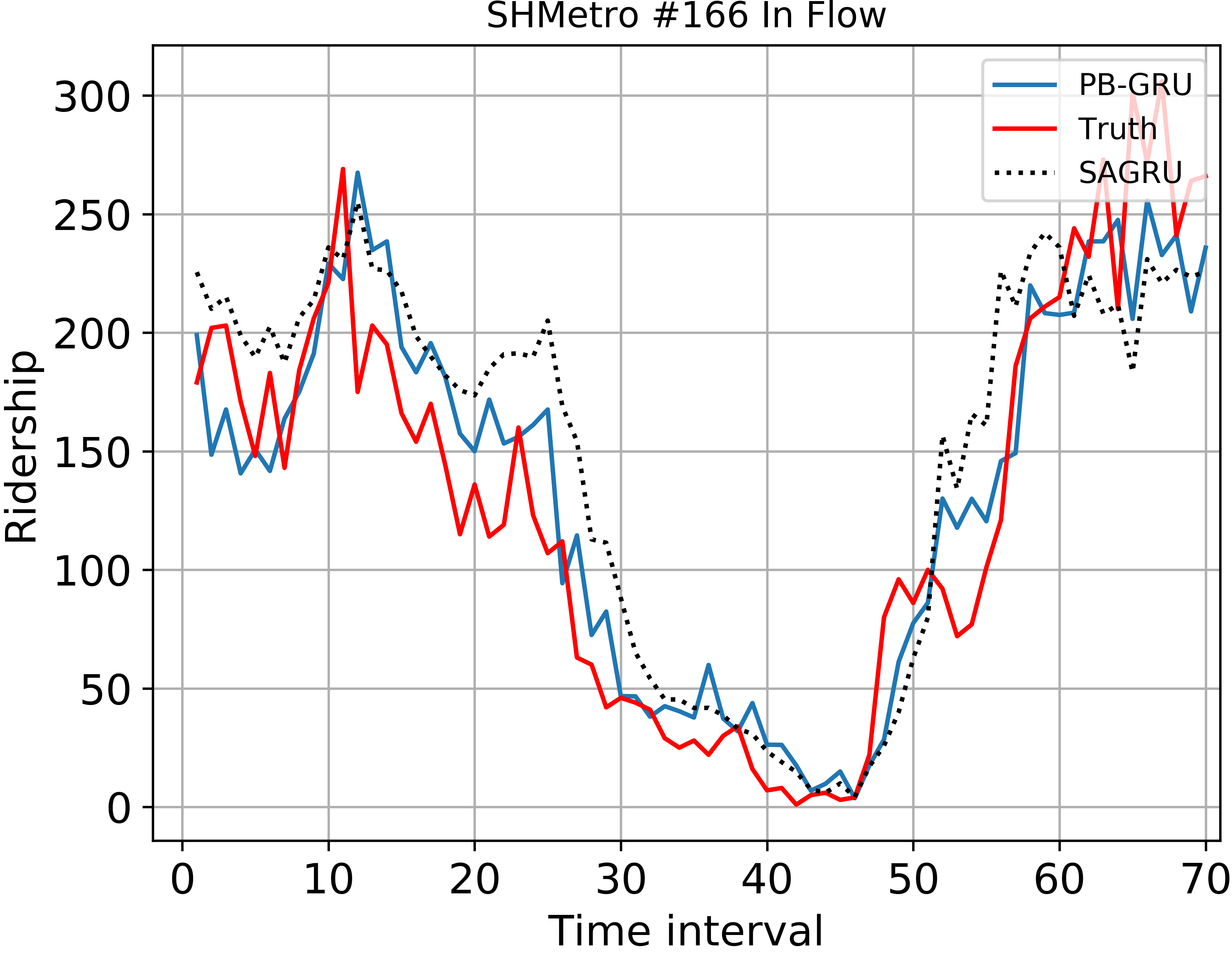}
}
\subfigure{
\includegraphics[width=8cm]{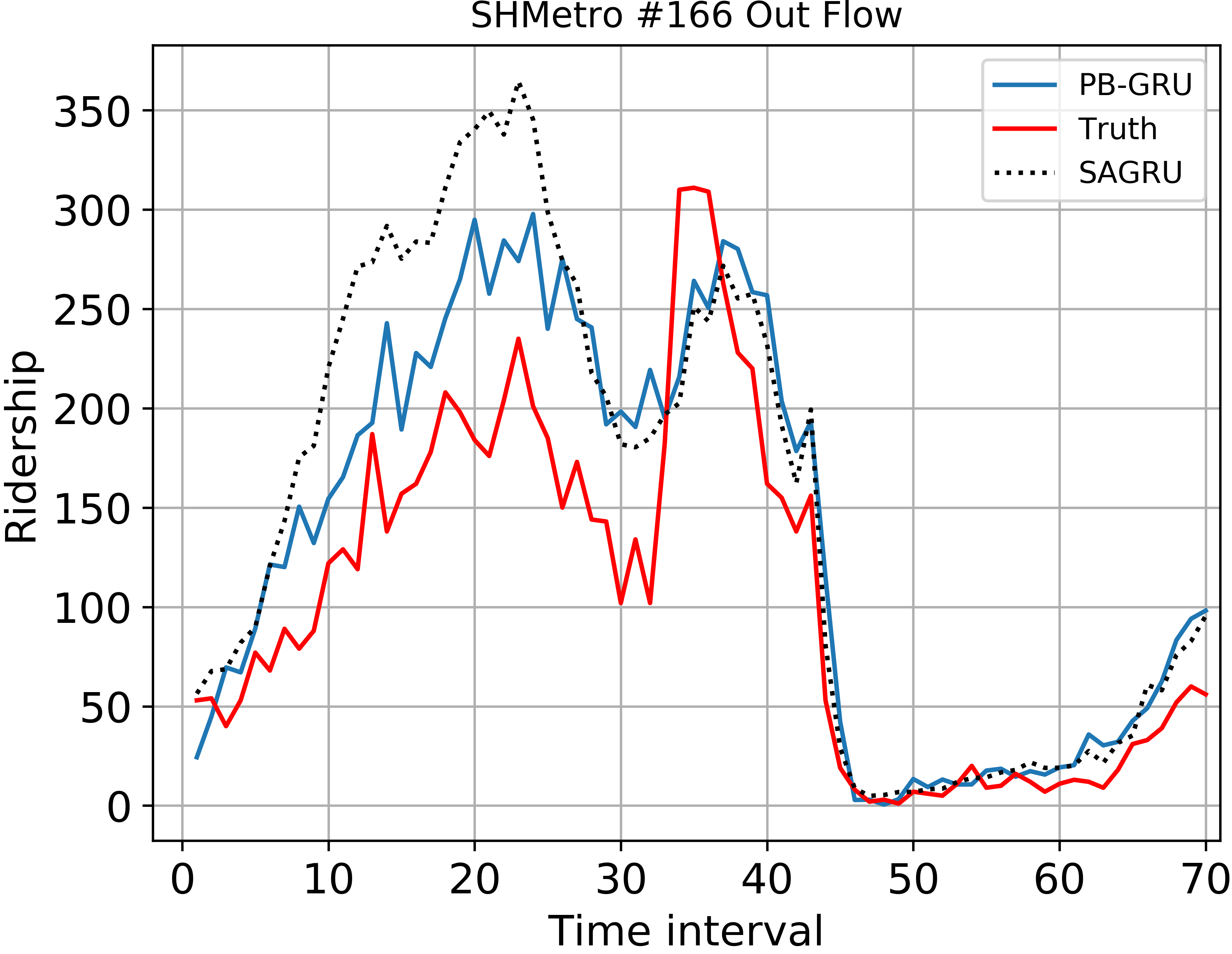}
}
\caption{Snapshot of three prediction instances. Station \#277 and \#12 are the lowest and highest average passenger volume in SHMetro respectively, Station \#166 is a more general instance with a different passenger volume pattern.}
\end{figure*}
error is obtained in different predicted interval tasks. It is very consistent with the common sense that the longer the interval is, the larger the travel range of passengers will be. It effectively proves that FDGCN can improve the prediction accuracy by modeling the inflow of the metro station within the K-hop. However, it also should be noted that the inappropriate K value has a certain negative impact on the prediction. Due to the differences between subway and urban planning, how to determine the value of K in different metro systems remain an open problem.

Finally, the training time of different methods is further compared on the same workstation. Table VI shows the time required to train an epoch for each method. Specifically, PVGCN and GCN-SUBLSTM, which are recently proposed for passenger volume prediction, are engaged in the comparison. We can see that PVGCN requires the longest time to train. This is because the computations of GRU cannot be paralleled, PVGCN integrates graph convolution into GRU's gating mechanism, and three graph convolution operations are performed for each forward propagation. PB-GRU and GCN-SUBLSTM require similar training time. It is worthy pointing out that our model has less training time. It is due to the fact that both FSGCN and FDGCN use SGC as graph convolution, which reduces a large number of parameters compared with the traditional GCN. In order to prove the efficiency of temporal attention, 
the additional training time of SAGRU was provided in the table. The attention in the model could be regarded as a gate mechanism with temporal complexity O(n), so increasing the size of training set will not lead to the exponential level increase of training time.
\begin{table}[H]
\setlength{\tabcolsep}{7mm}{
\renewcommand{\arraystretch}{1.3}
 \captionsetup{labelformat=default,labelsep=period}
\caption{ COMPARISON OF TRAINING TIME PER EPOCH}
\label{tab2}
\centering
\begin{tabular}{ccc}
\hline
Model& HZMetro& SHMetro\\
\hline
PVGCN&190.13s& 345.64s\\
\hline
SAGRU&1.05s&1.78s\\
\hline
GCN-SUBLSTM&1.44s&2.83s\\
\hline
PB-GRU&1.27s&2.19s\\
\hline
\end{tabular}}
\end{table}
\subsection{Visualization}
We visualized the prediction results of SHMetro to better demonstrate the model performance. Three instances are selected to illustrate the performance of PB-GRU on different passenger volumes. As shown in Fig. 7, PB-GRU is accurate in fitting both the overall trend and the peaks. With the help of FDGCN, the model is less likely to underestimate the out flow peak. This is an important reason why our model is lower than PVGCN and GCN-SUBLSTM on RMSE. Besides, we found that the periodicity of passenger flow in some subway stations (e.g.  \#166 station) is quite different. These special stations are major source of prediction error. However,  thanks to FSGCN and FDGCN, our method is able to handle such cases according to the visualization of station \#166.
\section{Conclusion And Future Works}
In this paper, a deep learning model PB-GRU is proposed for metro passenger volume prediction. In the model, We design two completely different graph convolution modules to capture flow pattern similarity and OD flow direction respectively. Based on the experiments on two real-world datasets, the proposed PB-GRU is superior to the existing models in terms of precision and training efficiency. More ablation studies further verify the effectiveness of the two graph convolution modules. In the future, we would further consider the variation of the passenger flow pattern within the daily period and the functions of different stations, so as to achieve more accurate prediction in specific city area.


\begin{thebibliography}{00}
\bibitem{b1} C. Ding, D. Wang, X. Ma, and H. Li, "Predicting short-term subway ridership and prioritizing its influential factors using gradient boosting decision trees," Sustainability, vol. 8, pp. 1100, 2016.
\bibitem{b2} W. Jiang and J. Luo, "Graph Neural Network for Traffic Forecasting: A Survey," arXiv preprint arXiv:2101.11174, 2021.
\bibitem{b3} L. Zhao, Y. Song, C. Zhang, Y. Liu, P. Wang, T. Lin, et al., "T-gcn: A temporal graph convolutional network for traffic prediction," IEEE Transactions on Intelligent Transportation Systems, vol. 21, pp. 3848-3858, 2019.
\bibitem{b4} Y. Li, R. Yu, C. Shahabi, and Y. Liu, "Diffusion Convolutional Recurrent Neural Network: Data-Driven Traffic Forecasting," in International Conference on Learning Representations, 2018.
\bibitem{b5} B. Yu, H. Yin, and Z. Zhu, "Spatio-temporal graph convolutional networks: a deep learning framework for traffic forecasting," in Proceedings of the 27th International Joint Conference on Artificial Intelligence, 2018, pp. 3634-3640.
\bibitem{b6} Z. Wu, S. Pan, G. Long, J. Jiang, and C. Zhang, "Graph WaveNet for deep spatial-temporal graph modeling," in International Joint Conference on Artificial Intelligence 2019, 2019, pp. 1907-1913.
\bibitem{b7} L. Liu, J. Chen, H. Wu, J. Zhen, G. Li, and L. Lin, "Physical-Virtual Collaboration Modeling for Intra-and Inter-Station Metro Ridership Prediction," IEEE Transactions on Intelligent Transportation Systems, 2020.
\bibitem{b8} X. Ma, J. Zhang, B. Du, C. Ding, and L. Sun, "Parallel architecture of convolutional bi-directional LSTM neural networks for network-wide metro ridership prediction," IEEE Transactions on Intelligent Transportation Systems, vol. 20, pp. 2278-2288, 2018.
\bibitem{b9} P. Chen, X. Fu, and X. Wang, "A Graph Convolutional Stacked Bidirectional Unidirectional-LSTM Neural Network for Metro Ridership Prediction," IEEE Transactions on Intelligent Transportation Systems, 2021.
\bibitem{b10} T. N. Kipf and M. Welling, "Semi-supervised classification with graph convolutional networks," arXiv preprint arXiv:1609.02907, 2016.
\bibitem{b11} F. Wu, A. Souza, T. Zhang, C. Fifty, T. Yu, and K. Weinberger, "Simplifying graph convolutional networks," in International conference on machine learning, 2019, pp. 6861-6871.
\bibitem{b12} X. He, K. Deng, X. Wang, Y. Li, Y. Zhang, and M. Wang, "Lightgcn: Simplifying and powering graph convolution network for recommendation," in Proceedings of the 43rd International ACM SIGIR Conference on Research and Development in Information Retrieval, 2020, pp. 639-648.
\bibitem{b13} Z. Cui, R. Ke, Z. Pu, and Y. Wang, "Deep bidirectional and unidirectional LSTM recurrent neural network for network-wide traffic speed prediction," arXiv preprint arXiv:1801.02143, 2018.
\bibitem{b14} J. Ye, J. Zhao, K. Ye, and C. Xu, "How to build a graph-based deep learning architecture in traffic domain: A survey," IEEE Transactions on Intelligent Transportation Systems, 2020.
\bibitem{b15} W. Jiang and J. Luo, "Big Data for Traffic Estimation and Prediction: A Survey of Data and Tools," arXiv preprint arXiv:2103.11824, 2021.
\bibitem{b16} T. Rakthanmanon, B. Campana, A. Mueen, G. Batista, B. Westover, Q. Zhu, et al., "Searching and mining trillions of time series subsequences under dynamic time warping," in Proceedings of the 18th ACM SIGKDD international conference on Knowledge discovery and data mining, 2012, pp. 262-270.
\bibitem{b17} A. Miglani and N. Kumar, "Deep learning models for traffic flow prediction in autonomous vehicles: A review, solutions, and challenges," Vehicular Communications, vol. 20, p. 100184, 2019.
\bibitem{b18} H. Yao, F. Wu, J. Ke, X. Tang, Y. Jia, S. Lu, et al., "Deep multi-view spatial-temporal network for taxi demand prediction," in Proceedings of the AAAI Conference on Artificial Intelligence, 2018.
\bibitem{b19} I. Sutskever, O. Vinyals, and Q. V. Le, "Sequence to sequence learning with neural networks," arXiv preprint arXiv:1409.3215, 2014.
\bibitem{b20} Z. Cui, R. Ke, Z. Pu, and Y. Wang, "Deep bidirectional and unidirectional LSTM recurrent neural network for network-wide traffic speed prediction," arXiv preprint arXiv:1801.02143, 2018.
\bibitem{b21} J. Zhang, Y. Zheng, and D. Qi, "Deep spatio-temporal residual networks for citywide crowd flows prediction," in Proceedings of the AAAI Conference on Artificial Intelligence, 2017.
\bibitem{b22} X. Shi, Z. Chen, H. Wang, D.-Y. Yeung, W.-K. Wong, and W.-c. Woo, "Convolutional LSTM network: A machine learning approach for precipitation nowcasting," Advances in neural information processing systems, vol. 28, 2015.
\bibitem{b23} W. Chen, L. Chen, Y. Xie, W. Cao, Y. Gao, and X. Feng, "Multi-Range Attentive Bicomponent Graph Convolutional Network for Traffic Forecasting," Proceedings of the AAAI Conference on Artificial Intelligence, vol. 34, pp. 3529-3536, 2020.
\bibitem{b24} C. Zheng, X. Fan, C. Wang, and J. Qi, "Gman: A graph multi-attention network for traffic prediction," in Proceedings of the AAAI Conference on Artificial Intelligence, pp. 1234-1241, 2020,
\bibitem{b25} C. Song, Y. Lin, S. Guo, and H. Wan, "Spatial-temporal synchronous graph convolutional networks: A new framework for spatial-temporal network data forecasting," in Proceedings of the AAAI Conference on Artificial Intelligence, 2020, pp. 914-921.
\bibitem{b26} J. Klicpera, A. Bojchevski, and S. Günnemann, "Predict then propagate: Graph neural networks meet personalized pagerank," arXiv preprint arXiv:1810.05997, 2018.
\bibitem{b27} M. Liu, H. Gao, and S. Ji, "Towards deeper graph neural networks," in Proceedings of the 26th ACM SIGKDD International Conference on Knowledge Discovery \& Data Mining, 2020, pp. 338-348.
\bibitem{b28} H. Zhu and P. Koniusz, "Simple spectral graph convolution," in International Conference on Learning Representations, 2021.
\bibitem{b29} Z. Liu, H. Chen, R. Feng, S. Wu, S. Ji, B. Yang, et al., "Deep Dual Consecutive Network for Human Pose Estimation," in Proceedings of the IEEE/CVF Conference on Computer Vision and Pattern Recognition, 2021, pp. 525-534.
\end{thebibliography}
\end{document}